\documentclass[10pt,twocolumn,letterpaper]{article}

\usepackage{cvpr}
\usepackage{times}
\usepackage{epsfig}
\usepackage{graphicx}
\usepackage{amsmath}
\usepackage{amssymb}
\usepackage{comment}
\def\vector#1{\mbox{\boldmath $#1$}}
\DeclareMathOperator*{\argmax}{arg\,max}

\cvprfinalcopy 

\def\cvprPaperID{****} 

\begin{document}

\title{MPM: Joint Representation of Motion and Position Map for Cell Tracking}

\author{Junya Hayashida~~~~~Kazuya Nishimura~~~~~Ryoma Bise\\
Kyushu University, Fukuoka, Japan
, {\tt\small \{bise@ait.kyushu-u.ac.jp\}}
}


\maketitle

\begin{abstract}
Conventional cell tracking methods detect multiple cells in each frame (detection) and then associate the detection results in successive time-frames (association).
Most cell tracking methods perform the association task independently from the detection task. However, there is no guarantee of preserving coherence between these tasks, and lack of coherence may adversely affect tracking performance.
In this paper, we propose the Motion and Position Map (MPM) that jointly represents both detection and association for not only migration but also cell division.
It guarantees coherence such that if a cell is detected, the corresponding motion flow can always be obtained.
It is a simple but powerful method for multi-object tracking in dense environments.
We compared the proposed method with current tracking methods under various conditions in real biological images and found that it outperformed the state-of-the-art (+5.2\% improvement compared to the second-best).
\end{abstract}

\vspace{-2mm}
\section{Introduction}
\vspace{-1mm}

Cell behavior analysis is an important research topic in the fields of biology and medicine. Individual cells must be detected and tracked in populations to analyze cell behavior metrics including cell migration speed and frequency of cell division. However, it is too time consuming to manually track a large number of cells. Therefore, automatic cell tracking is required.
Cell tracking in phase-contrast microscopy is a challenging multi-object tracking task. 
This is because the cells tend to have very similar appearance and their shapes severely deform. Therefore, it is difficult to identify the same cell among frames based on only its appearance. 
Moreover, cells in contact with each other often have blurry intercellular boundaries. 
In this case, it is difficult to identify touching cells from only one image. 
Finally, a cell may divide into two cells (cell mitosis), which is a very different situation from that of general object tracking.

The tracking-by-detection method is the most popular tracking paradigm because of the good quality of the detection algorithms that use convolutional neural networks (CNNs) for the detection step.
These methods detect cells in each frame and then associate the detection results among the time-frames by maximizing the sum of association scores of hypotheses, where the association step is often performed independently from the detection step~\cite{kanadeT2011,AlkofahiO2006,bise2011reliable}.
They basically use hand-crafted association scores based on the proximity and similarity in shape of cells, but the scores only evaluate the similarity of individual cells; {\it i.e.,} they do not use the context of nearby cells.
Moreover, the scores depend on the conditions of the time-lapse images, such as the frame interval, cell type, and cell density.

On the other hand, the positional relationship of nearby cells is important to identify the association when the density is high.
Several methods have been proposed for using context information~\cite{payer2018instance, hayashida2019cell}.
Hayashida {\it et al.}~\cite{hayashida2019cell} proposed a cell motion field (CMF) that represents the cell association between successive frames in order to directly predict cell motion (association) by using a CNN. This enables cell tracking at low frame rates.
This method outperformed the previous methods that use hand-crafted association scores.
However, it still independently performs the detection and association inference. This raises the possibility of incoherence between the two steps; {\it e.g.,} although a cell is detected, there is no response at the detected position in the CMF, and this would affect the tracking performance.

\begin{figure*}[t]
\begin{center}
   \includegraphics[width=0.93\linewidth]{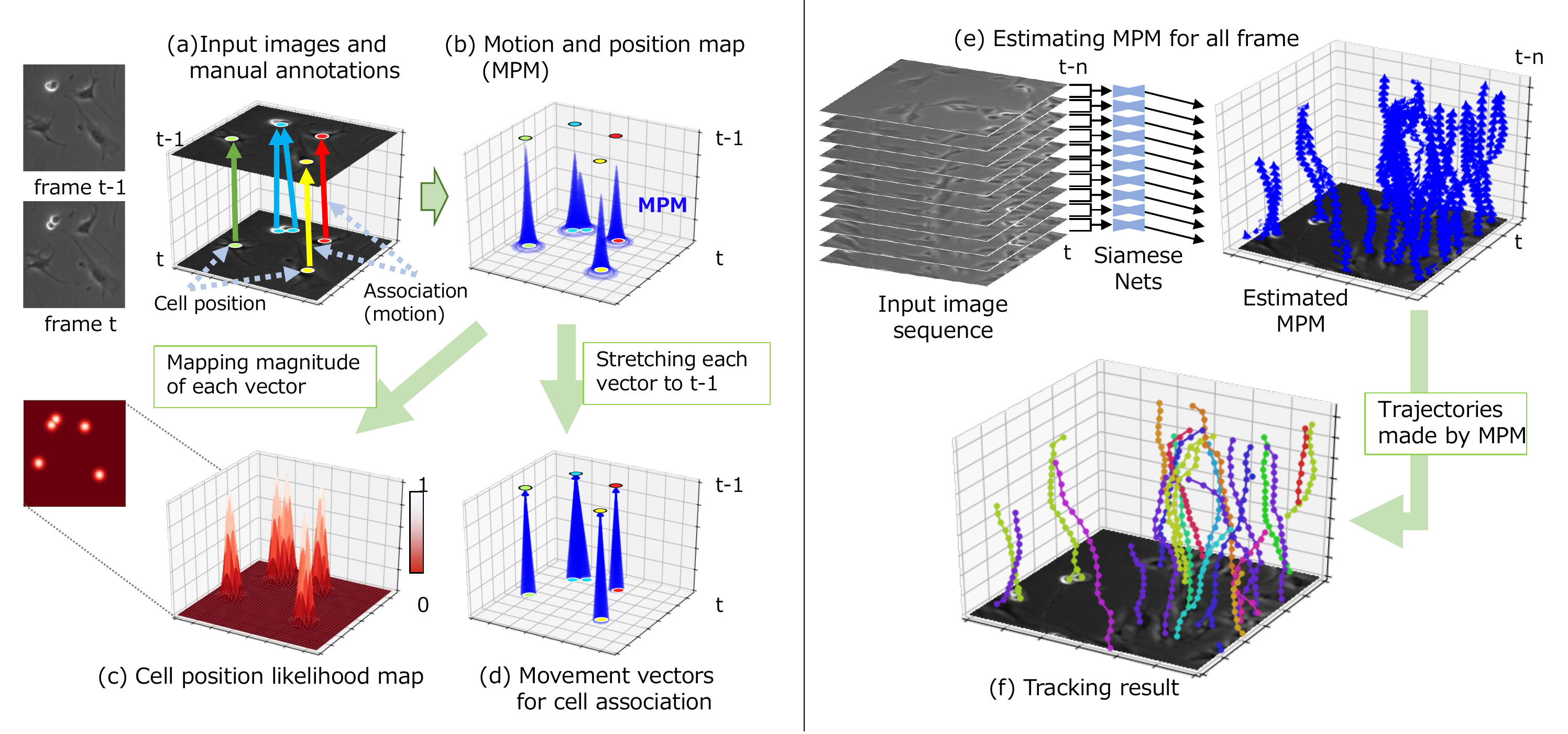}
\end{center}
\vspace{-3mm}
   \caption{Overview of the proposed method. (a) Input images in two successive frames and manual annotations of cell positions in each frame and their association. (b) Motion and position map (MPM) that jointly represents both detection and association. MPM encodes the following two pieces of information: (c) a cell-position likelihood map, which is a magnitude map of MPM. The peaks of this map indicate the cells' positions; (d) motion vector for a cell from $t$ to $t-1$. (e) Estimating the MPM for all frames by using siamese MPM-Nets. (f) The overall tracking results when using MPM.
  }
\label{fig:OVERALL}
\vspace{-3mm}
\end{figure*}

Many applications use multi-task learning~\cite{GkioxariG2014,MisraI2016,EigenD2015,ZengZ2019}, {\it i.e.,} learning multiple-related tasks.
This method basically has common layers express task-shared features and output branches that represent task-specific features. Multi-task learning improves the accuracy of the tasks compared with individual learning.
However, it is not easy to preserve the coherence between the detection and association tasks in multi-task learning.
If we simply apply a branch network, there is no guarantee that the coherence between these tasks will be preserved: {\it e.g.,} the network may detect a cell but not produce a corresponding association for the cell, which would adversely affect tracking performance. 

Unlike such multi-task learning methods, we propose the Motion and Position Map (MPM) that jointly represents both detection and association for not only migration but also cell division, as shown in Fig. \ref{fig:OVERALL}, where the distribution of magnitudes of MPM indicates the cell-position likelihood map at frame $t$; the direction of the 3D vector encoded on a pixel in MPM indicates the motion direction from the pixel at $t$ to the cell centroid at $t-1$.
Since MPM represents both detection and association, it guarantees coherence; {\it i.e.,} if a cell is detected, an association of the cell is always produced.
In addition, we can obtain the cell positions and their association for all frames by using siamese MPM networks, as shown in Fig. \ref{fig:OVERALL}(e). 
Accordingly, even a very simple tracking method can be used to construct overall cell trajectories with high accuracy (Fig. \ref{fig:OVERALL}(f)).
In the experiments, our method outperformed the state-of-the-arts under various conditions in real biological images.

Our main contributions are summarized as follows:
\begin{itemize}
    \item We propose the motion and position map (MPM) that jointly represents detection and association for not only migration but also cell division. It guarantees coherence such that if a cell is detected, the corresponding motion flow can always be obtained. It is a simple but powerful method for multi-object tracking in dense environments.
    \item By applying MPM-Net to the entire image sequence, we can obtain the detection and association information for the entire sequence. By using the estimation results, we can easily obtain the overall cell trajectories without any complex process.
\end{itemize}

\section{Related work}
\noindent {\bf Cell tracking:}
Cell tracking methods can be roughly classified into two groups: those with model-based evolution and those based on detection-and-association.
The first group uses sequential Bayesian filters such as particle filters~\cite{okuma2004boosted, smal2006bayesian}, or active contour models~\cite{li2008cell, wang2007cell, yang2005cell, zhou2019joint}. They update a probability map or energy function of cell regions by using the tracking results of the previous frame as the initial state for the updates.
These methods work well if cells are sparsely distributed.
However, these methods may get confused when cells are close together with blurry intercellular boundaries or move over a long distance.

The second group of cell-tracking methods are based on detection and association: first they detect cells in each frame; then they determine associations between successive frames.
Many cell detection/segmentation methods have been proposed to detect individual cells, by using level set~\cite{liu2010automated}, watershed~\cite{mkrtchyan2011efficient}, graph-cut~\cite{BenschR2015}, optimization using intensity features~\cite{bise2015}, physics-based-restoration~\cite{yin2012understanding, su2013cell,KangL2009},
weakly-supervised~\cite{nishimura2019weakly} and CNN-based detection methods~\cite{rempfler2017cell, rempfler2018tracing, akram2016joint,lux2019dic,arbelle2019microscopy}.
Moreover, many optimization methods have been proposed to associate the detected cells. They maximize an association score function, including linear programming for frame-by-frame association~\cite{kanadeT2011,BiseR2009,bise2013,zhou2019joint,WuZ12} and graph-based optimization for global data association~\cite{bise2011reliable, schiegg2013conservation, su2013cell,Funke18}.
These methods basically use the proximity and shape similarity of cells for making hand-crafted association scores. These association scores depend on the cell culture conditions, such as the frame interval, cell types and cell density.
In addition, the scores only evaluate the similarity of individual cells, not the context of nearby cells.
On the other hand, a human expert empirically uses the positional relationship of nearby cells in dense cases.
Several methods have been proposed on how to use such context information~\cite{payer2018instance, hayashida2019cell,rempfler2017cell}.
Payer {\it et al.}~\cite{payer2018instance} proposed a recurrent stacked hourglass network (ConvGRU) that not only extracts local features, but also memorizes inter-frame information. 
However, this method requires annotations for all cell regions as training data, and it does not perform well when the cells are small.
Hayashida {\it et al.}~\cite{hayashida2019cell} proposed a cell motion field that represents the cell association between successive frames that can be estimated with a CNN. Their method is a point-level tracking, and thus it dramatically reduces annotation costs compared with pixel-level cell boundary annotation.
These methods have been shown to outperform those that use hand-crafted association scores.
However, the method still performs the detection and association inference independently. This causes incoherence between steps, which in turn affects tracking performance.

\noindent {\bf Object tracking for general object:}
For single-object tracking that finds the bounding box corresponding to given initial template, Siamese networks have been often used to estimate the similarity of semantic features between the template and target images~\cite{Valmadre2017SiamFC,LiB2018SiamRPN}.
To obtain a good similarity map, this approach is based on representation learning. However, in the cell tracking problem, there may be many cells that have similar appearances; thus, the similarity map would have a low confidence level.
Tracking-by-detection (detection-and-association) is also a standard approach in multi-object tracking (MOT) for general objects. 
In the association step, features of objects and motion predictions~\cite{WangL2017,ZhaoD2018} are used to compute a similarity/distance score between pairs of detection and/or tracklets.
Re-identification is often used for feature representation~\cite{YuF2016, SadeghianA2017}. This method does not use the global spatial context, and the detection and association processes are separated.
Optimization techniques~\cite{XiaoW2016, HenschelR2019} have been proposed for jointly performing the detection and association processes; they jointly find the optimal solution of detection and association from a candidate set of detections and associations.
Sun {\it et al.}~\cite{SunS2019} proposed the Deep Affinity Network (DAN). Although this method jointly learns a feature representation for identification and association, the detection step is independent of these learnings. 

\noindent {\bf Multi-task learning:}
Multi task learning (MTL) has been widely used in computer vision for learning similar tasks such as pose estimation and action recognition ~\cite{GkioxariG2014}; depth prediction and semantic classification \cite{MisraI2016, EigenD2015}; and for room floor plans~\cite{ZengZ2019}.
Most multi-task learning network architectures for computer vision are based on existing CNN architectures~\cite{MisraI2016,DoerschC2017,KokkinosI2017}.
These architectures basically have common layers and output branches that express both task-shared and task-specific features.
For example, cross-stitch networks~\cite{MisraI2016} contain one standard CNN per task, with cross-stitch units to allow features to be shared across tasks.
Such architectures require a large number of network parameters and scale linearly with the number of tasks. In addition, there is no guarantee of coherence among the tasks if we simply apply the network architecture to a multi-task problem such as detection and association. This means that we need to design a new loss function or architecture to preserve the coherence depending on the pairs of the tasks.
To the best of our knowledge, multi-task learning methods specific to our problem have not been proposed.

Unlike these methods, we propose MPM for jointly representing the detection and association tasks at once. The MPM can be estimated using simple U-net architecture, and it guarantees coherence such that if a cell is detected, the corresponding motion flow can always be obtained.

\section{Motion and Position Map (MPM)}
Our tracking method estimates the Motion and Position Map (MPM) with a CNN.
The MPM jointly represents the position and moving direction of each cell between successive frames by storing a 3D vector on a 2D plane.
From the distribution of the 3D vectors in the map, we can obtain the positions of the cells and their motion information.
First, the distribution of the magnitudes of the vectors in the MPM shows the likelihoods of the cell positions in frame $t$.
Second, the direction of the 3D vector on a cell indicates the motion direction from the pixel at $t$ to the cell centroid at $t-1$.
Note that we treat the cell motion from $t$ to $t-1$ in order to naturally define a cell division event when a mother cell divides into two daughter cells by one motion from a cell.
By using MPM as the ground-truth, it is possible to use a simple CNN model that has a U-net architecture, that we call MPM-Net, to perform the simultaneous cell detection and association tasks.

\begin{figure}[t]
\begin{center}
   \includegraphics[width=\linewidth]{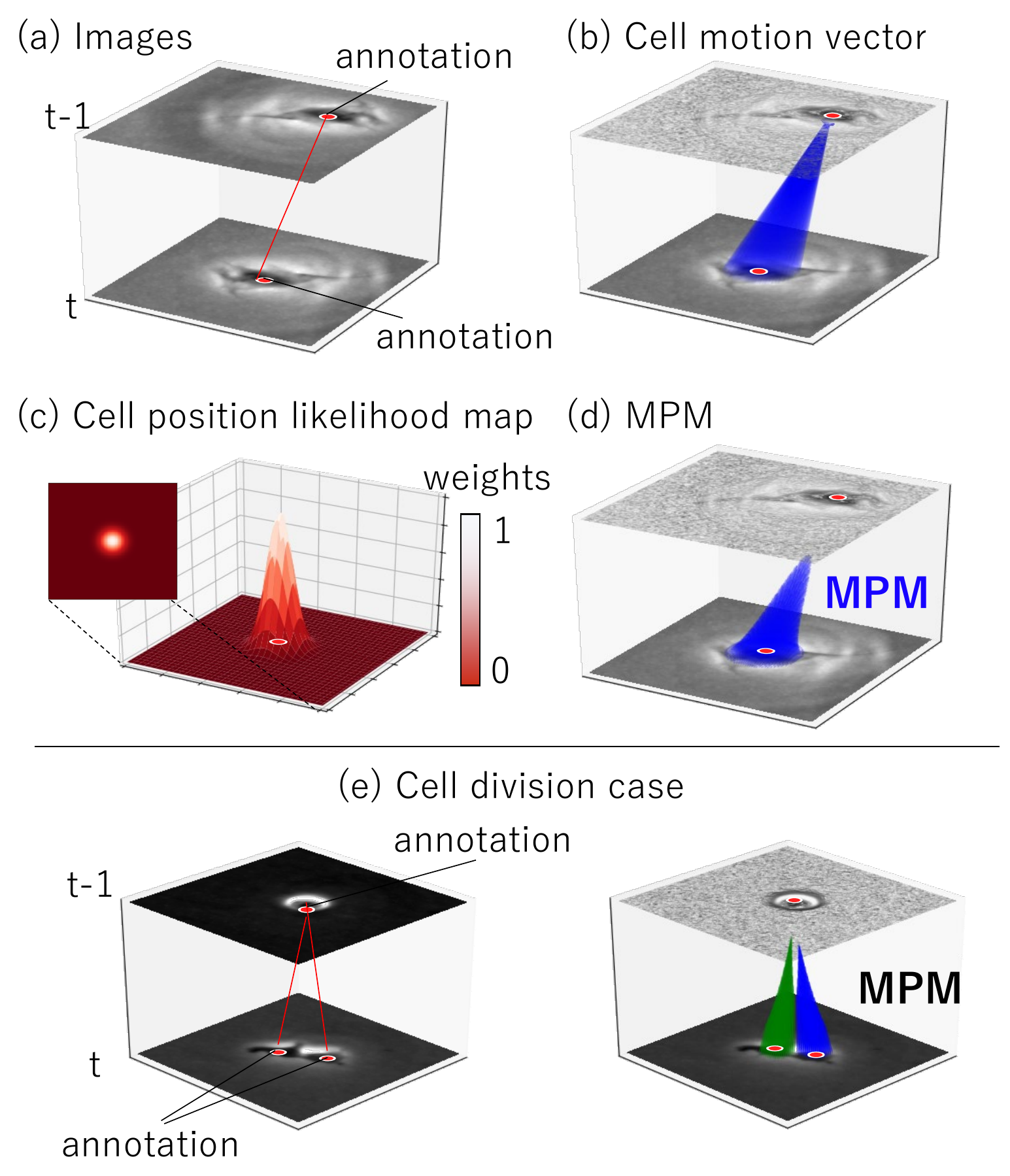}
\end{center}
   \caption{Example of the generated MPM from the given annotated cell positions at $t$ and $t-1$. Note that this example shows the enlarged images for a single cell for explanation. (a) Input images with the annotation points. The red points indicate the annotated points on the rough centroids of cells. (b) Motion vectors from the neighbor points around the annotated point at $t$ to the annotated point $t-1$. (c) Cell position likelihood map that indicates the weight map of the vector of MPM. (d) MPM that represents both the motion vector and the position likelihood map, where it encodes the 3D vector at each pixel and the magnitude of the vector indicates the cell position likelihood, the direction of the vector indicates the direction of the motion. (e) Example of the cell division case, where two cells are associated to the single cell.}
\label{fig:mpm}
\end{figure}

First, we explain how to generate the ground truth of MPM from the manual annotation.
Fig. \ref{fig:mpm} shows the simple example of the annotations for the motion of a single cell and it's MPM.
The manual annotation for an image sequence expresses the position of each cell position in terms of 3D orthogonal coordinates by adding a frame index to the annotated 2D coordinate, where the same cell has the same ID from frame to frame.
We denote the annotated cell positions for cell $i$ at frame $t-1$ and $t$ as $\mathbf{a}^{i}_{t-1}$, $\mathbf{a}^{i}_{t}$ respectively. They are defined as:
\begin{align}
\begin{cases}\mathbf{a}_{t-1}^{i} &= (x_{t-1}^i, y_{t-1}^i, t-1)^{\rm T},\\
\mathbf{a}_{t}^{i} &= (x_t^i, y_t^i, t)^{\rm T}.
\end{cases}\label{eq:position}
\end{align}

Although the annotated position for a cell is a single point at $t$, manual annotation may not always correspond to the true position.
Similar to \cite{hayashida2019cell}, we make a likelihood map of cell positions, where an annotated cell position becomes a peak and the value gradually decreases in accordance with a Gaussian distribution, as shown in Fig. \ref{fig:mpm}.
Similarly, the motion vectors are represented as the 3D vectors pointing to $\mathbf{a}_{t-1}^{i}$ from the nearby pixels of $\mathbf{a}_{t}^{i}$. 

To generate ground truth MPM $\mathbf{C}_{t-1,t}$, we first generate individual MPM $\mathbf{C}_{t-1,t}^{i}$ for each cell $i$ in frame $t-1$ and $t$.
Each individual MPM is obtained by storing 3D vectors pointing to $\mathbf{a}_{t-1}^{i}$ from the neighboring region of $\mathbf{a}_{t}^{i}$.
Let $\vector{v}_{t-1,t}^i=\mathbf{a}_{t-1}^i-\mathbf{p}_t$ be the motion vector from each coordinate $\mathbf{p}_{t} = (x_t, y_t, t)^{\rm T}$ at frame $t$ to the annotated position $\mathbf{a}_{t-1}^i$. $\mathbf{C}_{t-1,t}^{i}$ is defined as:
\begin{equation}
\mathbf{C}_{t-1,t}^{i} (\mathbf{p}_t, \vector{v}_{t-1,t}^i) = w(\mathbf{p}_t)\frac{\vector{v}_{t-1,t}^{i}}{||\vector{v}_{t-1,t}^{i}||_2},\label{eq:indiC}
\end{equation}
where $w(\cdot)$ is a weight function that takes each coordinate at frame t as input and returns a scalar value. 
Since the motion vector $\vector{v}_{t-1,t}^i/||\vector{v}_{t-1,t}^i||_2$ is a unit vector in Eq.\eqref{eq:indiC}, $w(\cdot)$ is a function that represents the magnitude of each vector in MPM.
$w(\cdot)$ is a Gaussian function with the annotation coordinate $\mathbf{a}_{t}^{i}$ as the peak, and is defined as:
\begin{equation}
w(\mathbf{p}_t^i) = \operatorname{exp}\left (-\frac{||\mathbf{a}_{t}^{i}-\mathbf{p}_{t}||_2^2}{\sigma^2}\right),
\label{eq:w}
\end{equation}
where $\sigma$ is a hyper-parameter that indicates the standard deviation of the distribution and controls the spread of the peak.

The vector stored at each coordinate of MPM $\mathbf{C}_{t-1,t}$ is the vector of the individual MPMs with the largest value of the coordinates among all the magnitudes of the individual MPMs.
The MPM $\mathbf{C}_{t-1, t}$ that aggregates all the individual MPM is defined as:
\begin{align}
\label{eq:maxMPM}
\mathbf{C}_{t-1,t} (\mathbf{p}_t) &= \mathbf{C}_{t-1,t}^{\hat{i}}(\mathbf{p}_t),\\
\hat{i} &= \argmax_{i}||\mathbf{C}_{t-1,t}^{i}(\mathbf{p}_t)||.
\end{align}
Fig. \ref{fig:mpm}(d) shows an example of the MPM generated based on the annotation (a).
One of the advantage of the MPM is that can represent the cell division case that two daughter cells are associated to a mother cell, as shown in Fig. \ref{fig:mpm}(e).

A U-Net is trained with the ground truth of the MPM by using a loss function that is the mean squared error (MSE) between the training data and the estimated data, where two images are treated as two channels of the input, and the outputs of a pixel are 3 channels of the 3D vector.
Let the ground-truth of MPM be $\mathbf{c}$ and the estimated MPM be $\hat{\mathbf{c}}$. The loss function is defined as:
\begin{equation}
\mathbf{Loss}(\mathbf{c},\mathbf{\hat{c}})=\frac{1}{n}\sum_{i=1}^n(||\mathbf{c_i}-\mathbf{\hat{c_i}}||_2^2 +({||\mathbf{c_i}||_2}-||\mathbf{\hat{c_i}}||_2)^2)\label{eq:loss}
\end{equation}
where the first term $||\mathbf{c_i}-\mathbf{\hat{c_i}}||_2^2$ is a square error between $\mathbf{c}$ and $\hat{\mathbf{c}}$, and the second term $({||\mathbf{c_i}||_2}-||\mathbf{\hat{c_i}}||_2)^2$ is a square error between the magnitudes of $c$ and $\hat c$. 
Here, we empirically added the second term that directly reflects the error of the magnitude
(i.e., detection). Even though we use only the first term, it works, but the proposed loss function was stable.

\begin{figure}[t]
\begin{center}
   \includegraphics[width=\linewidth]{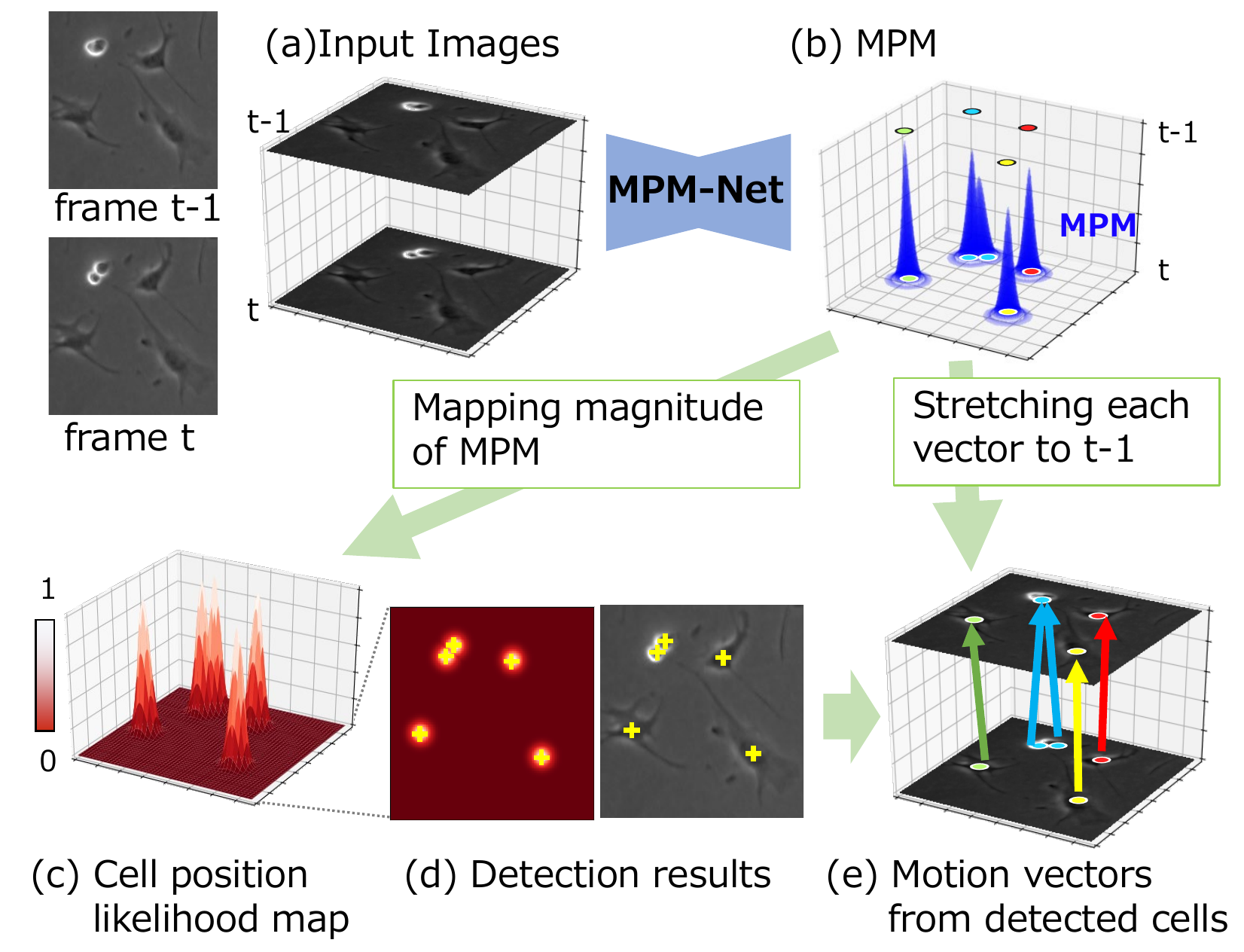}
\vspace{-4mm}
\end{center}
   \caption{Overview of cell detection and association in inference. (a) Input images. (b) MPM estimated by MPM-Net. (c) Cell-position likelihood map determined from the magnitude of MPM. (d) The peaks are detected as cells at $t$. (e) The cell motion from $t$ to $t-1$ is estimated by the direction vector at the detected peak point of MPM.}
\label{fig:inference_overall}
\vspace{-1mm}
\end{figure}

\section{Cell tracking by MPM}
Fig. \ref{fig:inference_overall} shows an overview of cell detection and association in inference. 
For frame $t$ and $t-1$ (Fig. \ref{fig:inference_overall}(a)), we estimate the MPM $\mathbf{C}_{t-1,t}$ by using the trained MPM-Net (Fig. \ref{fig:inference_overall}(b)).
The positions of cells in frame $t$ are detected from the magnitude of $\mathbf{C}_{t-1,t}$ (Fig. \ref{fig:inference_overall}(c)(d)), and the cells in frame $t-1$ are associated with each 3D vector stored at the detected positions (Fig. \ref{fig:inference_overall}(e)).

\subsection{Cell detection by the magnitude of MPM}
The cell-position likelihood map $\mathbf{L}_{t-1,t}$ at frame $t$ is obtained from the distribution of magnitudes of MPM $\mathbf{C}_{t-1,t}$.
The detected cell positions are the coordinates of the each local maximum of $\mathbf{L}_{t-1,t}$. 
In the implementation, we applied Gaussian smoothing as a pre-processing against noise to $L_{t-1,t}$.

\subsection{Cell association by MPM}
\begin{figure}[t]
\begin{center}
   \includegraphics[width=0.67\linewidth]{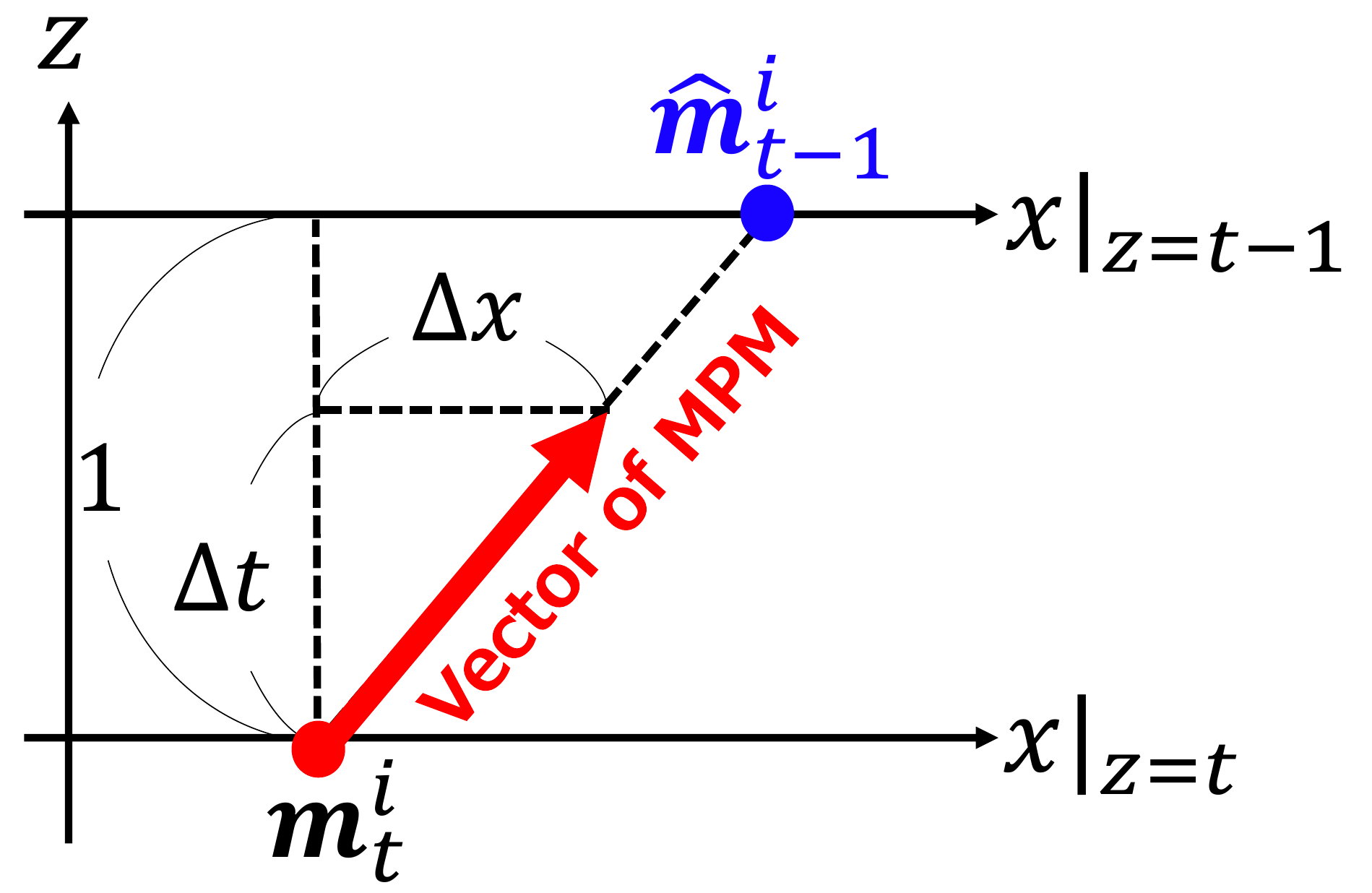}
\end{center}
\vspace{-1mm}
   \caption{Estimation of the cell position at frame $t-1$}
\label{fig:triangle}
\vspace{-1mm}
\end{figure}

The direction of each 3D vector of MPM $\mathbf{C}_{t-1,t}$ indicates the direction from position of the cell at $t$ to position of the same cell at $t-1$.
The position at frame $t-1$ is estimated using the triangle ratio from each vector of $\mathbf{C}_{t-1,t}$ stored at each detected position in frame $t$, as shown in Fig. \ref{fig:triangle}.
Let the detected position of each cell $i$ be $\vector{m}_t^{i}=(x_t^i,y_t^i,t)^{\rm T}$, and the 3D vector stored in $\mathbf{m}^i_t$ be $\vector{v}_{t-1,t}^i=(\Delta x,\Delta y,\Delta t)^{\rm T}$. The estimated position $\hat{\mathbf{m}}^{i}_{t-1}$ for each cell $i$ at frame $t-1$ is defined as:
\begin{equation}
\hat{\mathbf{m}}_{t-1}^i=(x_t^i+\frac{\Delta x}{\Delta t},y_t^i+\frac{\Delta y}{\Delta t},t-1)^{\rm T}.
\label{eq:xv}
\end{equation}

\begin{figure}[t]
\begin{center}
   \includegraphics[width=0.9\linewidth]{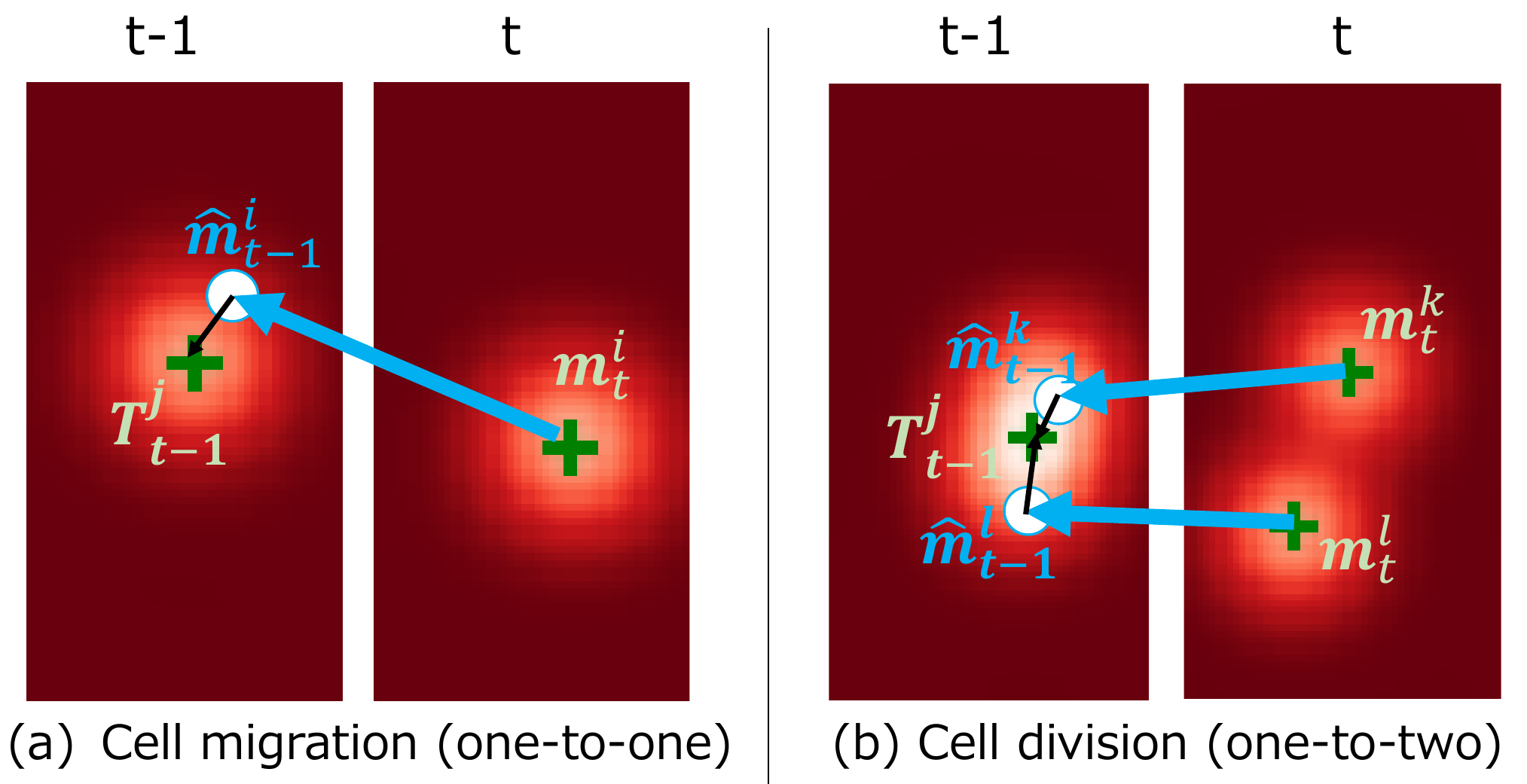}
\end{center}
\vspace{-1mm}
   \caption{Illustration of association process. (a) The case of cell migration (one-to-one matching): the estimated position $\mathbf{m}_{t-1}^i$ is associated with the nearest tracked cell $\mathbf{T}_{t-1}^j$. (b) The case of cell division (one-to-two matching): two cells $\mathbf{m}_{t-1}^l$, $\mathbf{m}_{t-1}^k$ are associated as daughter cells of $\mathbf{T}_{t-1}^j$.}
\label{fig:association}
\vspace{-1mm}
\end{figure}

We associate the detected cells in frame $t$ with the tracked cells in frame $t-1$ on the basis of $\hat{\mathbf{m}}_{t-1}^i$ and the previous confidence map $\mathbf{L}_{t-2,t-1}$ used for the cell detection in frame $t-1$, as shown in Fig. \ref{fig:association}.
The estimated position $\hat{\mathbf{m}}_{t-1}^i$ of cell $i$ at $t-1$ may be different from the tracked point at $t-1$ which is located at a local maximum of cell-position likelihood map.
Thus, we update the estimated position $\mathbf{m}_{t-1}^i$ to a certain local maximum that is the position of a tracked cell $\mathbf{T}_{i-1}^j$ by using the hill-climbing algorithm to find the peak (the position of the tracked cell).
Here, the $j$-th cell trajectory from $s_j$ until $t-1$ is denoted as $\mathbf{T}^j=\{\mathbf{T}^j_{s_j},...,\mathbf{T}^j_{t-1}\}$, where $s_j$ indicates the frame in which the $j$-th cell trajectory started.
If $\hat{\mathbf{m}}_{t-1}^i$ was updated to a position with $\mathbf{T}_{i-1}^j$, we associate $\mathbf{m}_{t}^i$ with $\mathbf{T}_{t-1}^j$ as the tracked cell $\mathbf{T}_{t}^j=\mathbf{m}_{t}^i$. If the confidence value at the estimated position at $t-1$ is zero ({\it i.e.,} the point is not associated with any cell), $\mathbf{m}_{t-1}^i$ is registered as a newly appearing cell $\mathbf{T}^{N_{t-1}+1}$, where $N_{t-1}$ is the number of cell trajectories until time $t$ before the update.

If two estimated cell positions $\mathbf{m}_{t-1}^k$, $\mathbf{m}_{t-1}^l$ were updated to one tracked cell position $\mathbf{T}_{t-1}^j$, these detected cells are registered as new born cells $\mathbf{T}^{N+1}=\{\mathbf{m}_{t-1}^k\}$, $\mathbf{T}^{N+2}=\{\mathbf{m}_{t-1}^l\}$, {\it i.e.,} daughter cells of $\mathbf{T}^j$, and the division information is also registered $\mathbf{T}^j \rightarrow \mathbf{T}^{N_{t-1}+1},\mathbf{T}^{N_{t-1}+2}$.
The cell tracking method using MPM can be performed very simply for both cell migration and division.

\subsection{Interpolation for undetected cells}
The estimated MPM at $t$ may have a few false negatives because the estimation performance is not perfect.
In this case, the trajectory $\mathbf{T}^j$ is not associated with any cell and the frame-by-frame tracking for the cell is terminated at $t-1$ for the moment.
The tracklet $\mathbf{T}^j$ is registered as a temporarily track-terminated cell.
During the tracking process using MPM, if there are newly detected cells including daughter cells from cell division at $t-1+q$, the method tries to associate the temporarily track-terminated cell using MPM again, where the MPM is estimated by inputting the images at $t-1$ and $t-1+q$ ({\it i.e.,} the time interval of the images is $q$).
Next, the cell position is estimated and updated by using frame-by-frame association.
If the newly detected cell is associated with one of the temporarily track-terminated cells, the cell is then associated with the cell and the positions in the interval of frames, in which corresponding cells were not detected, are interpolated based on the vector of the MPM.
If there are multiple temporarily track-terminated cells, this process is iterated on all track-terminated cells in ascending order of time.
The method excludes a cell from the list of the track-terminated cells when the interval from the endpoint of the track-terminated cell to the current frame is larger than a threshold.
Here, we should note that cells at $t$ tend to be associated with another cell as cell divisions when a false negative occurs at $t-1$. 
To avoid such false-positives of cell-division detection, the method tries to associate such daughter cells with the track-terminated cells. This cell tracking algorithm can be performed online.

\begin{figure}[t]
\begin{center}
   \includegraphics[width=0.75\linewidth]{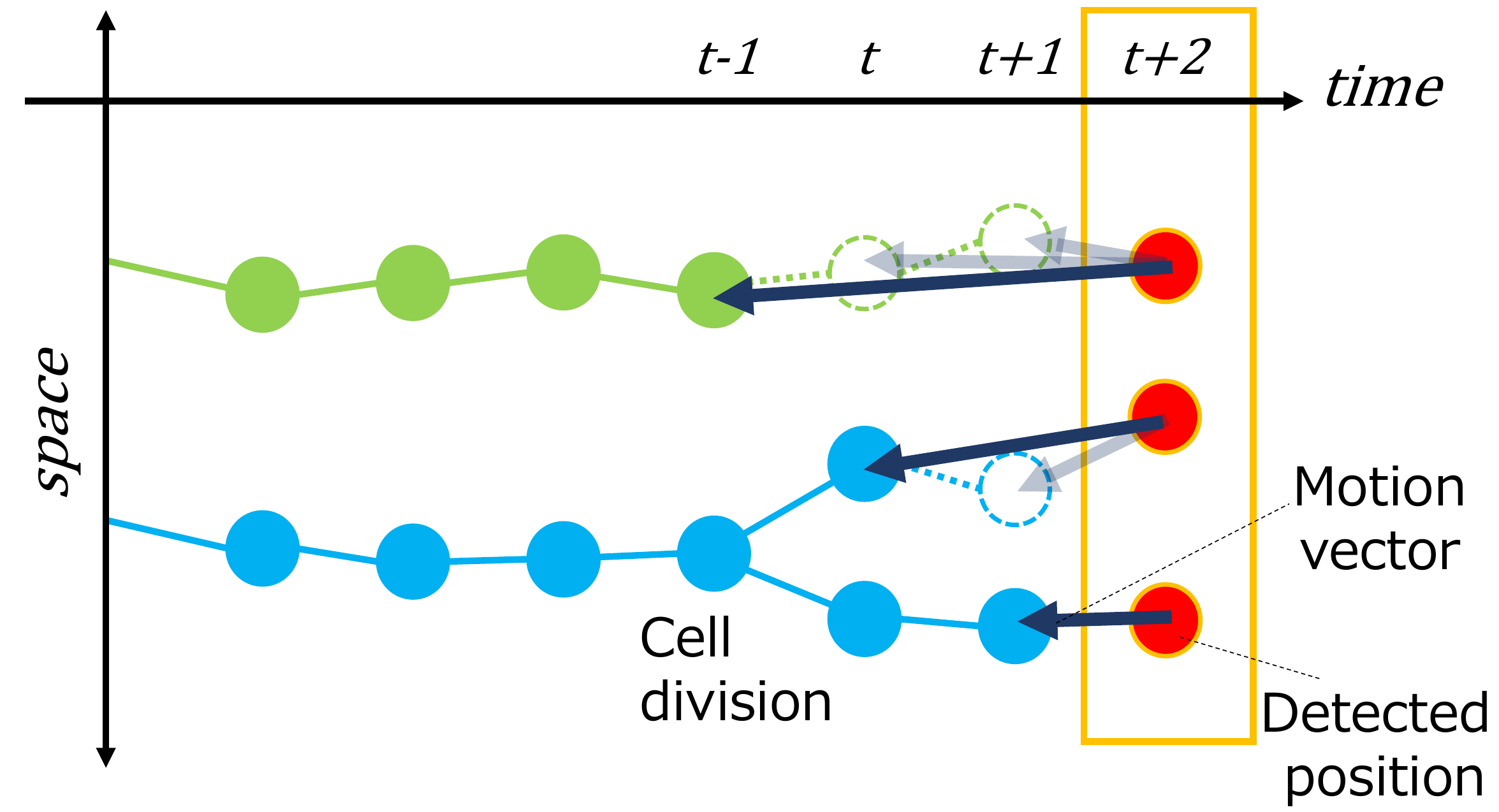}
\vspace{-2mm}
\end{center}
   \caption{Illustration of interpolation process for undetected cells. The newly detected cells (red) are associated with the track-terminated cells (green) using MPM.}
\label{fig:long}
\vspace{-2mm}
\end{figure}

Fig. \ref{fig:long} shows an example of this process, in which the tracking of the green trajectory is terminated at $t-1$ due to false negatives at $t$, $t+1$, and the red cell is detected at $t+2$. In this case, the MPM is estimated for three frame intervals (the inputs are the images at $t-1$ and $t+2$), and it is associated with the track-terminated cell. Then, the cell positions at $t$ and $t+1$ are interpolated using the vector of MPM and the tracking starts again.


\section{Experiments}

\begin{figure}[t]
\centering
    \includegraphics[width=\linewidth]{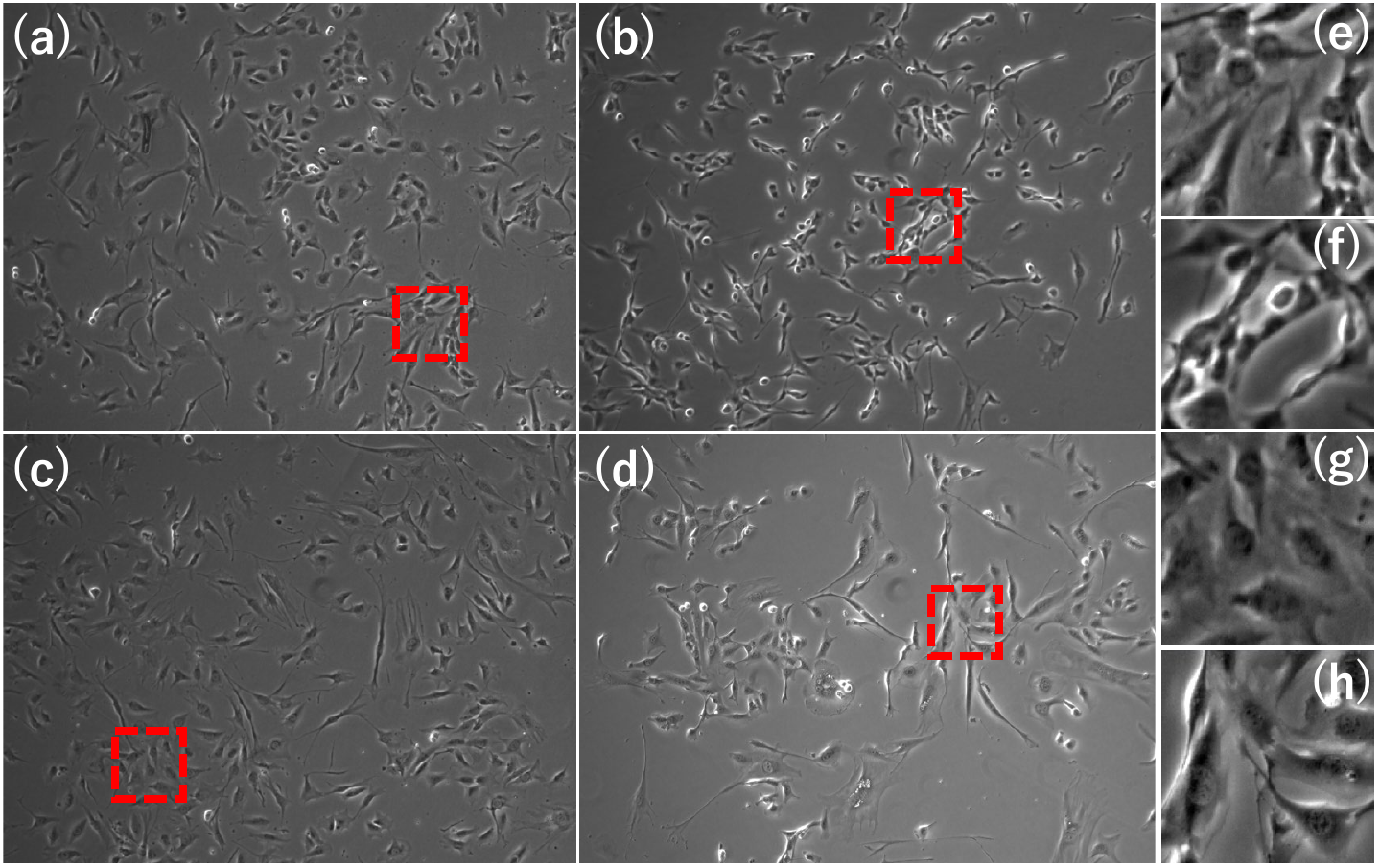}
    \caption{Example images from four conditions. a) Control, b) FGF2, c) BMP2, d) BMP2+FGF2, (e)-(h) the enlarged images of the red box in (a)-(d), respectively. The appearance of cells are different depending on the culture conditions. Cells often shrink and partially overlapped in FGF2, cells tend to be spread under BMP2, there are both cells who are spread or shrunk under BMP2+FGF2.}
    \label{fig:cellconditions}
    \vspace{-4mm}
\end{figure}

\subsection{Data set}
In the experiments, we used the open dataset~\cite{elmerK2018} that includes time-lapse image sequences captured by phase-contrast microscopy since this is the same setup with our target to track. In this data-set, the point-level ground truth is given, in which the rough cell centroid was annotated with cell ID, and hundreds of mouse myoblast stem cells were cultured under four cell culture conditions that are different growth-factor mediums; a) Control (no growth factor), b) FGF2, c) BMP2, d) BMP2+FGF2, where there are four image sequences for each condition (total: 16 sequences).
Since a stem cell is differentiated depending on the growth factor, the cell appearance changes over the four culture conditions as shown in Fig. \ref{fig:cellconditions}.

The images were captured every 5 minutes and each sequence consists of 780 images with the resolution of 1392$\times$1040 pixels.
All cells were annotated in a sequence in BMP2, and three cells were randomly picked up at the beginning of the sequence, and then the three cell's family trees through the 780-th image were annotated for all the other 15 sequences, where the number of the annotated cells increased with time due to cell division. The total number of annotated cells in the 16 sequences is 135859~\cite{kanadeT2011}.

Since our method requires the training data that all cells should be fully annotated in each frame, we additionally annotated cells in a part of frames under three conditions; Control: 100 frames, BMP2: 100 frames, FGF2+BMP2: 200 frames.
Totally the annotated 400 frames were used as the training data, and the rest of the data was used as ground truth in the test.
To show the robustness for the cross-domain data (the different culture conditions from training data) in the test, we did not make the annotation for FGF2.
In all experiments, we set the interval value $q$ for interpolation as 5.

\subsection{Performance of cell Detection}
The detection performance of our method compared with the three other methods including the state-of-the-art; Bensch~\cite{BenschR2015} that is based-on graph-cut, Bise~\cite{bise2011reliable} that is based on physics model of phase-contrast microscopy~\cite{yin2012understanding}, and Hayashida~\cite{hayashida2019cell} (the state-of-the-art for this data-set) that is based on deep learning that estimates the cell-position likelihood map that is similar to ours. Here, the annotation of this dataset is the point-level annotation, and thus we could not apply the learning-based segmentation methods that require the pixel-level annotation for the individual cell regions.
We used the fully-annotated sequence to measure the precision, recall, and F1-score as the detection performance metrics. For the training and tuning the parameters, we used the same training sequences. 

\begin{figure}[t]
\centering
    \includegraphics[width=\linewidth]{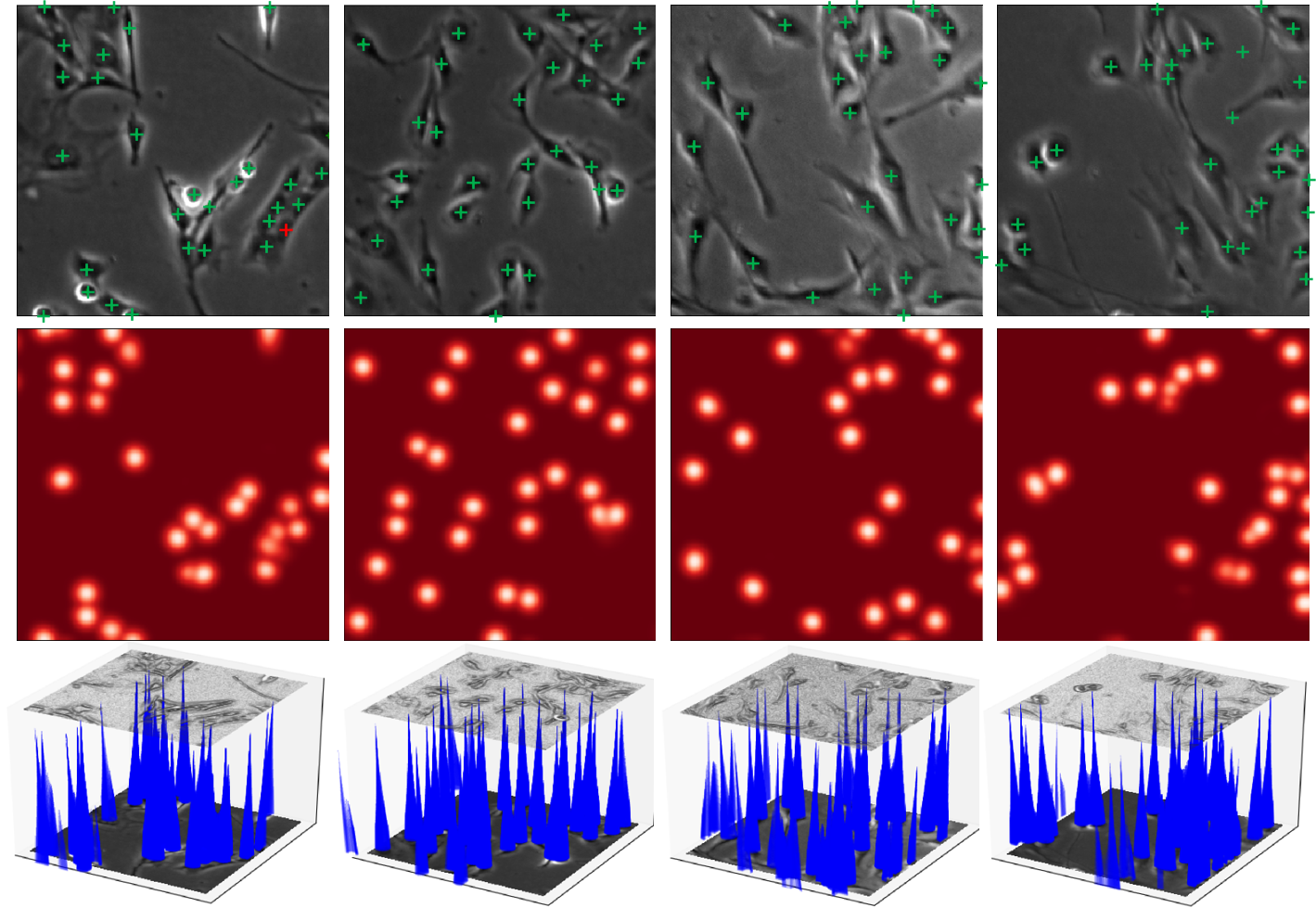}
    \caption{Examples of our cell detection results. Top: detection results. Green '+' indicates the true positive, red '+' indicates the false negative. Middle: cell position likelihood map generated from the estimated MPM (Bottom).}
    \label{fig:detectionresults}
    \vspace{-2mm}
\end{figure}

Fig. \ref{fig:detectionresults} shows examples of our cell detection results. Our method successfully detected the cells that have various appearances, including spread cells, small brighter cells, and touching cells.
Table \ref{Tab:CellDetection} shows the results of the cell detection evaluation. The deep learning-based methods (Hayashida's and ours) outperformed the other two methods. The detection method in \cite{hayashida2019cell} estimated the cell-position likelihood map by U-net, in which their method represents the detection but not for the association. The terms of recall and F1-score of our method are slightly better (+1.4\%) than those of their method.
Since our method uses the context from two successive frames in contrast to their method that detects cells each frame independently, we consider that it would be possible to improve the performance of detection.

\subsection{Performance of cell Tracking}
Next, we evaluated the tracking performance of our method with four other methods; Bensch~\cite{BenschR2015} based-on frame-by-frame association with graph-cut segmentation; Chalfoun~\cite{ChalfounJ2016} based-on optimization for the frame-by-frame association with the segmentation results by \cite{yin2012understanding}; Bise\cite{bise2011reliable} based on spatial-temporal global data association; Hayashida~\cite{hayashida2019cell} based on motion flow estimation by CNN.
Hayashida's method is the most related to the proposed method. Their method estimates the cell position map and motion flow by CNN, independently.

\begin{table}[t]
\caption{Cell detection performance on the terms of precision, recall and F1-score.}
\label{Tab:CellDetection}
\begin{tabular}{l|cccc}
Method&\itshape \begin{tabular}{c}Bensh\\\cite{BenschR2015}\end{tabular}&\itshape \begin{tabular}{c}Bise\\\cite{bise2011reliable}\end{tabular}&\itshape \begin{tabular}{c}Hayashida\\\cite{hayashida2019cell}\end{tabular}&\itshape Ours \\\hline\hline
precision & 0.583  & 0.850 & \textbf{0.968} & 0.964\\
recall    & 0.623  & 0.811 & 0.902 & \textbf{0.932}\\
F1-score  & 0.602 & 0.830 & 0.934 & \textbf{0.948}\\
\end{tabular}
\end{table}

\begin{table}[t]
\caption{Association Accuracy.}
\label{Tab:AA}
\begin{tabular}{l|lllll}
Method & \hspace{-1mm}Cont.\hspace{-1mm} & \hspace{-1mm}FGF2\hspace{-1mm} & \hspace{-1mm}BMP2\hspace{-1mm} & \hspace{-3mm}\begin{tabular}{l}FGF2+\\BMP2\end{tabular}\hspace{-3mm} & \hspace{-1mm}Ave.\\\hline\hline
\itshape Bensh\cite{BenschR2015}\hspace{-2mm}       &\hspace{-1mm} 0.604 & 0.499 & 0.801 & 0.689 &\hspace{-2mm} 0.648\\
\itshape Chalfoun\cite{ChalfounJ2016}\hspace{-2mm}               &\hspace{-1mm} 0.762 & 0.650 & 0.769 & 0.833 &\hspace{-2mm} 0.753\\
\itshape Bise\cite{bise2011reliable}\hspace{-2mm}   &\hspace{-1mm} 0.826 & 0.775 & 0.855 & 0.942 &\hspace{-2mm} 0.843\\
\itshape Hayashida\cite{hayashida2019cell}\hspace{-2mm}&\hspace{-1mm} 0.866 & 0.884 & 0.958 & 0.941 &\hspace{-2mm} 0.912\\
\itshape Ours (MPM)\hspace{-2mm}                    &\hspace{-1mm} \textbf{0.947} & \textbf{0.952} & \textbf{0.991} & \textbf{0.987} &\hspace{-2mm} \textbf{0.969}\\
\end{tabular}
\end{table}

\begin{table}[t]
\caption{Target Effectiveness.
}
\label{Tab:TE}
\begin{tabular}{l|lllll}
Method & \hspace{-1mm}Cont.\hspace{-1mm} & \hspace{-1mm}FGF2\hspace{-1mm} & \hspace{-1mm}BMP2\hspace{-1mm} & \hspace{-3mm}\begin{tabular}{l}FGF2+\\BMP2\end{tabular}\hspace{-3mm} & \hspace{-1mm}Ave.\\\hline\hline
\itshape Bensh\cite{BenschR2015}\hspace{-2mm}       &\hspace{-1mm} 0.543 & 0.448 & 0.621 & 0.465 &\hspace{-2mm} 0.519\\
\itshape Chalfoun\cite{ChalfounJ2016}\hspace{-2mm}               &\hspace{-1mm} 0.683 & 0.604 & 0.691 & 0.587 &\hspace{-2mm} 0.641\\
\itshape *Li\cite{KangL2009}\hspace{-2mm}          &\hspace{-1mm} 0.700 & 0.570 & 0.630 & 0.710 &\hspace{-2mm} 0.653\\
\itshape *Kanade\cite{kanadeT2011}\hspace{-2mm}    &\hspace{-1mm} \textbf{0.830} & 0.640 & 0.800 & 0.790 &\hspace{-2mm} 0.765\\
\itshape Bise\cite{bise2011reliable}\hspace{-2mm}   &\hspace{-1mm} 0.733 & 0.710 & 0.788 & 0.633 &\hspace{-2mm} 0.771\\
\itshape Hayashida\cite{hayashida2019cell}\hspace{-2mm}&\hspace{-1mm} 0.756 & 0.761 & 0.939 & 0.841 &\hspace{-2mm} 0.822\\
\itshape Ours (MPM)\hspace{-2mm}                    &\hspace{-1mm} 0.803 & \textbf{0.829} & \textbf{0.958} & \textbf{0.911} &\hspace{-2mm} \textbf{0.875}\\
\end{tabular}
\vspace{-2mm}
\end{table}

We used two quantitative criteria to assess the performance: association accuracy and target effectiveness \cite{kanadeT2011}.
To compute association accuracy, each target (human-annotated) was assigned to a track (computer-generated) for each frame. The association accuracy was computed as the number of true positive associations divided by the number of associations in the ground-truth. If a switch error occurred between two cells A and B, we count the two false positives (A$\rightarrow$~B, B$\rightarrow$~A).
To compute target effectiveness, we first assign each target to a track that contains the most observations from that ground-truth. Then target effectiveness is computed as the number of the assigned track observations over the total number of frames of the target. It indicates how many frames of targets are followed by computer-generated tracks. This metric is a stricter metric. If a switching error occurs in the middle of the trajectory, the target effectiveness is 0.5.

\begin{figure}[t]
\centering
    \includegraphics[width=\linewidth]{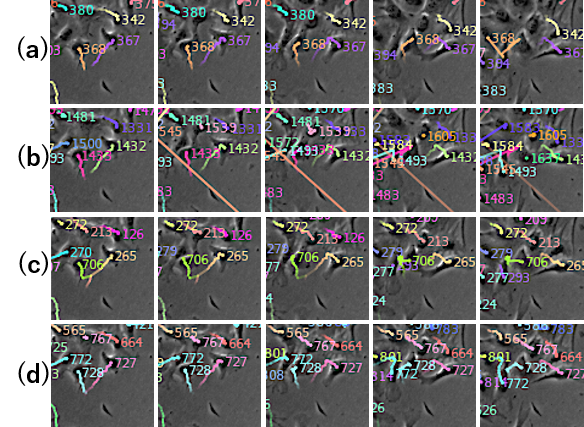}
    \caption{Tracking results from each compared method under BMP2. (a) Bise, (b) Chalfoun, (c) Hayashida, (d) ours. Horizontal axis indicate the time.}
    \label{fig:results}
    \vspace{-2mm}
\end{figure}

Table \ref{Tab:AA} and \ref{Tab:TE} show the performance of the association accuracy and the target effectiveness, respectively. In these tables, the average scores for each condition (three or four sequences in each condition) are denoted~\footnote{The scores of the target effectiveness of *Li and *Kanade were evaluated by the same data-set in their papers. Since the association accuracy of their methods \cite{KangL2009,kanadeT2011} were not described in their papers, we could not add their performance at the Table \ref{Tab:AA}.}.
In Table \ref{Tab:TE}, we show the additional two methods as the reference; Li~\cite{KangL2009} based on level-set tracking and Kande~\cite{kanadeT2011} based on frame-by-frame optimization. 
The performances of Bensh, Chalfoun, Li were sensitive depending on the culture conditions, and the performances were not good on both the metric terms, in particular, (FGF2) and (BMP2+FGF2). 
We consider that the sensitivity of their detection methods adversely affects the tracking performance.
The methods of Bise and Kanade that use the same detection method~\cite{yin2012understanding} achieved better accuracy compared to the three methods thanks to the better detection.
The state-of-the-art method (Hayashida) furthermore improved the performances (Fig. \ref{fig:results}(c)) compared with the other current methods by estimating cell motion flow and position map independently.
Our method (MPM) outperformed all the other methods (Fig. \ref{fig:results}(d)).
It improved +5.7\% of association accuracy and +5.1\% of the target-effectiveness on the average score compared to the second-best (Hayashida).

\begin{figure}[t]
\centering
    \includegraphics[width=\linewidth]{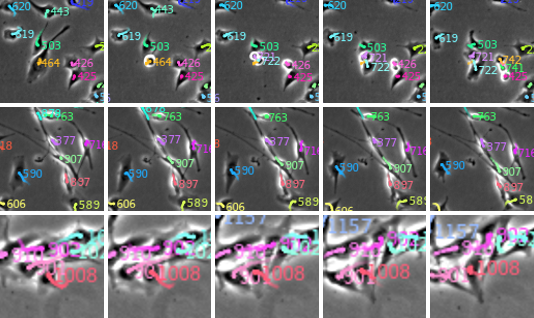}
    \caption{Tracking results from our method under each culture conditions; (a) Control, (b) FGF2, (c) BMP2+FGF2. Although the cell appearances are different depending on the conditions, our method correctly tracked the cells.}
    \label{fig:results2}
    \vspace{-2mm}
\end{figure}

\begin{figure}[t]
\centering
    \includegraphics[width=\linewidth]{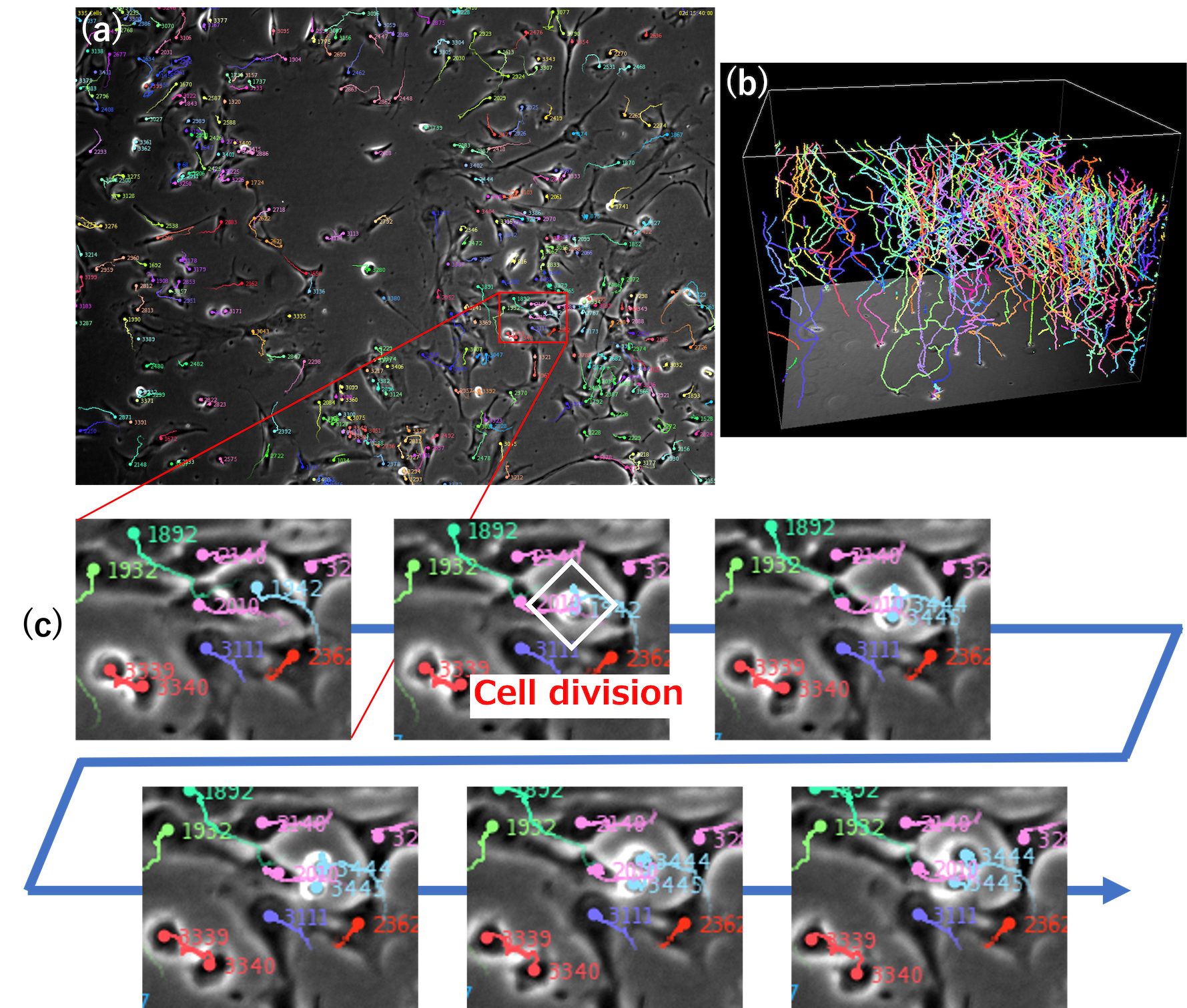}
    \caption{Examples of our tracking results. (a) Entire image. (b) 3D view of estimated cell trajectories. The z-axis indicates the time, and each color indicates the trajectory of a single cell. (c) Enlarged image sequence at the red box in (a), in which our method correctly identified the cell division and tracked all cells.}
    \label{fig:track}
    \vspace{-2mm}
\end{figure}

Fig. \ref{fig:results2} shows the tracking results from ours under each condition, in which our method correctly tracked the various appearance of cells. Fig. \ref{fig:track} shows the tracking results in the case when a cell divided into two cells in the enlarged image-sequence. Our method could correctly identify the cell division and track all cells in the region.

\section{Conclusion}
In this paper, we proposed the motion and position map (MPM) that jointly represents both detection and association, which can be represented in the cell division case not only migration.
The MPM encodes a 3D vector into each pixel, where the distribution of magnitudes of the vectors indicates the cell-position likelihood map, and the direction of the vector indicates the motion direction from the pixel at $t$ to the cell centroid at $t-1$.
Since MPM represents both detection and association, it guarantees coherence; {\it i.e.,} if a cell is detected, the association of the cell is always produced, and thus it improved the tracking performance.
In the experiments, our method outperformed the state-of-the-art method with over 5\% improvement of the tracking performance under various conditions in real biological images. In future work, we will develop an end-to-end tracking method that can estimate the entire cell trajectories from the entire sequence in order to use the global spatial-temporal information on CNN.

\section*{Acknowledgments}
This work was supported by JSPS KAKENHI Grant Numbers JP18H05104 and JP18H04738.

{\small
\bibliographystyle{ieee_fullname}
\bibliography{main_bib}
}
\end{document}